\documentclass[letterpaper, 10pt, conference, hyphens]{ieeeconf}      % Use this line for a4 paper

\let\labelindent\relax

\usepackage{amsmath}

\usepackage{amsthm,amssymb}
\usepackage[utf8]{inputenc}
\usepackage{algorithm}
\usepackage{algorithmic}
\usepackage{graphicx}
\usepackage{hyperref}
\usepackage{subfig}
\usepackage{multirow}
\usepackage{graphicx}

\DeclareMathOperator*{\argmax}{arg\,max} 
\usepackage{enumitem}
\usepackage{centernot}
\usepackage{lipsum}
\usepackage{blindtext}
\usepackage{scrextend}
\usepackage{stackengine}
\usepackage{tikz}
\usepackage{todonotes}
\usepackage{amsfonts}

\usepackage{gensymb}

\theoremstyle{definition}
\newtheorem{definition}{Definition}[section]
\usepackage{mathtools}
\usepackage[hyphenbreaks]{breakurl}
\newcommand{\ownership}[1][m]{\textit{#1-Ownership}}
\newcommand{\ownershipchange}{\textit{Ownership\_Change}}
\newcommand{\bhonwershipchange}[3]{\ownershipchange$[k, #1 \xrightarrow{#3, L} #2]$}
\newcommand{\occlusionaware}{\textit{Occlusion\_Aware}}
\newcommand{\planalg}{\texttt{ALG}}
\newcommand{\occlusionshare}{\textit{Occlusion\_Share}}

\IEEEoverridecommandlockouts                              % This command is only needed if 
\title{\LARGE \bf
Emergent Cooperative Behavior in Distributed Target Tracking with Unknown Occlusions
}

\author{Tianqi Li$^{1}$, Lucas W. Krakow$^2$ and Swaminathan Gopalswamy$^{1}$% <-this % stops a space
\thanks{$^{1}$Tianqi Li and Swaminathan Gopalswamy are with the Department of Mechanical Engineering, 
        Texas A\&M University, College Station, TX 77840, USA
        {\tt\scriptsize \{tianqili, sgopalswamy\}@tamu.edu}}%
\thanks{$^{2}$Lucas W. Krakow is with
the Bush Combat Development Complex, Texas A\&M University, Bryan, TX 77807, USA
\burl{lwkrakow@tamu.edu}}
\thanks{This
work was supported by Army Research Laboratories under contract W911NF1920243.
This work is maintained at
\burl{https://github.com/TianqiLi7398/emergent_behavior_active_tracking}
}
}

\begin{document}

\maketitle
\thispagestyle{empty}
\pagestyle{plain}

%%%%%%%%%%%%%%%%%%%%%%%%%%%%%%%%%%%%%%%%%%%%%%%%%%%%%%%%%%%%%%%%%%%%%%%%%%%%%%%%
\begin{abstract}
% The target tracking task generates the estimation of the object of interest (OOI) by recursively updating beliefs based on sensor information.
% However, the limit of sensors' field of view (FoV) and occlusion in FoV constrains the sensing input, and further, tracking performance.
% We demonstrate that in a multi-target tracking task, A multi-robot sensing fleet is able to overcome such occlusion influence in an unknown environment through:
% a) active motion planning to optimize tracking performance;
% b) distributed sensor fusion and cooperative planning schema;
% c) detecting and maintaining the occlusion in the environment.

% Alternate Abstract:

Tracking multiple moving objects of interest (OOI) with multiple robot systems (MRS) has been addressed by active sensing that maintains a shared belief of OOIs and plans the motion of robots to maximize the information quality. 
Mobility of robots enables the behavior of pursuing better visibility, which is constrained by sensor field of view (FoV) and occlusion objects.
We first extend prior work to detect, maintain and share occlusion information explicitly, allowing us to generate occlusion-aware planning even if \`{a} priori semantic occlusion information is unavailable. 
The efficacy of active sensing approaches is often evaluated according to estimation error and information gain metrics.
However, these metrics do not directly explain the level of cooperative behavior engendered by the active sensing algorithms. 
Next, we extract different emergent cooperative behaviors that stem from the same underlying algorithms but manifest differently under differing scenarios. 
In particular, we highlight and demonstrate three emergent behavior patterns in active sensing MRS: 
(i) Change of tracking responsibility between agents when tracking trajectories with divergent directions or due to a re-allocation of the resource among heterogeneous agents;
(ii) Awareness of occlusions to a trajectory and temporal leave-and-return of the sensing agent; 
(iii) Sharing of local occlusion objects in MRS that subsequently improves the awareness of occlusion.

\end{abstract}

\begin{keywords}
Active Sensing, Visibility, Occlusion-Aware, Cooperative Emergent Behavior, Dynamic Occlusion Map
\end{keywords}

\section{Introduction}

% As the practical applications for robotics continues to grow, the purpose of sensing has evolved from just providing information to the robot for its localization and navigation, to providing additional contextual information on the environment. 
% The ability to track different moving objects (\emph{targets}) in the environment using the sensors on robots is of great interest. 
% Tracking of a single target is often achieved by converting the raw measurements from the sensors to an observation of the target location using the sensor platform's location (which may itself be estimated), generating a classification of the target properties which leads to a dynamic motion model of the target, and finally combining the observations with the motion model using  standard estimation techniques such as the Kalman Filter to arrive at target location estimates. \ref{??}. When there are multiple targets that need to be tracked, then additional techniques to associate the observations with the maintained tracks is required. In such cases, techniques such as the Joint Probabilisitc Data Association (JPDA) is utilized. \ref{}

As the practical robotics applications continue to grow, the purpose of sensing has evolved from just providing information to the robot for its localization and navigation to providing additional contextual information on the environment. 
The ability to track different moving objects in the environment using robot sensors is of great interest. 
Several procedures typically accomplish tracking of targets: 
First, the raw sensor signal is processed by layers of object recognition, object localization, and classification; 
second, the observation data containing semantic and positional information is processed by data association algorithms that associate observation with estimations; 
third, the maintained trajectories get updated based on observations;
and finally track management system handles the conditions like trajectory identification, deletion, birth, etc. 

% Target tracking perceptual algorithms' focus

Tracking using the above methods becomes challenging when the object of interest (OOI) is not observable because the sensor's Field of View (FoV) is limited or some objects occlude the OOI. 
Prior studies have addressed this by focusing on improving a tracked object's estimation error while comprehending environmental occlusions.
For example, a visual odometry maintains robust depth estimation of moving targets given occlusions via a multi-scale global attention module \cite{lu2021global}.
The local context feature and global semantic information are integrated to moderate the interference of occlusion and deformation in a Siamese network visual tracking approach in \cite{yao2022semantic}.
Additional semantic information on the tracked targets and the environment is crucial in processing the sensed signals and associating the information with tracked targets.
The social force model for human motion enables the joint probabilistic data association filter (JPDAF) in human tracking scenarios with inter-agent interaction and occlusion.
Learned variational Bayesian clustering helps estimate the number of targets containing occluded ones in \cite{ur2015multi}. 
% The above approaches essentially rely upon prior semantic information on the location of occlusions so the estimation is improved.
% In this paper, an alternate approach is proposed where occlusion information is dynamically created, and allows use of target tracking even when prior semantic information on occlusions is not available.  

% Introduce planning's role in active sensing

While the approaches above only focus on the sensing part of target tracking, the robot motion itself can be planned to extract task-relevant (e.g., OOI) information. 
The study of \emph{active sensing / active target tracking} is inspired by the behavior of intelligent animals \cite{yang2016theoretical}. 
One implementation of an active sensing robot is a healthcare ground robot, which tracks people in a closed area of interest (AOI) \cite{bandyopadhyay2009motion}.
The planner for the active tracking is built based on a partially observable Markov decision process (POMDP). Rather than greedily maintaining OOIs inside its FoV, the robot allows the temporal absence of OOIs to balance power consumption and target visibility.
In another single robot case, a pursuit-evasion game for an observer-OOI pair demonstrates the optimal trajectory planning in space-time coordinates \cite{zou2018optimal}.

% from visibility side: occlusion and fov scenarios

The planning of active tracking can also be extended to consider occlusions. 
\textit{Visibility}-aware planning studies the chance of detection based on the state of the OOIs, occlusion objects, and ego state and configuration. 
Occlusion hinders sensing visibility typically in two patterns:
Horizontal opaque obstacles in the FoV of ground robots, for example, walls \cite{gemerek2022directional, jung2002tracking, zou2018optimal}, people \cite{bandyopadhyay2009motion} and neighboring vehicles \cite{buckman2020generating};
downside shadow of trees \cite{tallamraju2019active, li2021optimizing, li2022sma}, containers \cite{jakob2010occlusion}, or other robots
\cite{hausman2016occlusion} in the vertical FoV of unmanned aerial vehicles (UAV). 
The above approaches rely upon prior semantic information on the location of occlusions, so the estimation is improved.

Multiple robot systems can be used to jointly track the targets to address the increasing scope of the observation area and the number of targets. 
Extending active sensing from a single robot platform to multi-robot systems (MRS) has led to several interesting research studies, such as the study of active cooperative perception \cite{spaan2010active},
information-driven control \cite{ferrari2021information}, 
and subsequent applications in robotic rescue \cite{mobedi20113}, event recording and analysis \cite{wu2019interactive, shell2019planning},
surveillance and exploration \cite{kantaros2019asymptotically}, etc.
MRS algorithms can be extended to consider occlusions as in the single robot case. 
The impact of the size and density of tree occlusion on the distributed target tracking (DTT) is analyzed in \cite{li2022sma}.

% active sensing's task and areas

% semantic understanding of the map
% \cite{bernreiter2019multiple} environmental semantic understanding in SLAM.

% planning with cooperation

An MRS is not a simple aggregation of multiple single robot systems. 
The redundancy in the FOVs of the agents as they move over the operating region provides significant opportunities for collaboration and improvement in tracking. 
Explicit cooperative algorithms have been researched, such as the region-based (dividing AOI into FoVs of robots) and target-follow behavior patterns in visibility maximization in complex environments with occlusions \cite{jung2002tracking}. 
Also, the formation control idea, like target enclosing by redundant sensing robots, is presented in \cite{lopez2019adaptive}.

On the other hand, implicit algorithms that focus on a common minimization objective have also proven successful - the experiments \cite{tallamraju2019active}, and simulation \cite{li2022sma} of UAV team demonstrations with various tree occlusion scenarios.
The general look-ahead approaches such as POMDP and model predictive control (MPC) have been successfully applied in active sensing \cite{bandyopadhyay2009motion, buckman2020generating, li2021optimizing, li2022sma, miller2009coordinated}. 
While the objective function can measure the algorithm's efficacy, recognizing the cooperative behaviors is non-trivial. 
The behavior-level analysis is valuable as this enables the introduction of human reasoning to work in conjunction with algorithmic decision-making. 
Prior work in behavior extraction and analysis in the context of robots and vehicles includes:  
Definition of different reactive controllers from a behavioral perspective in Braitenberg vehicle model \cite{braitenberg1986vehicles}, taxonomy for behaviors in herds, schools, and swarms \cite{reynolds1987flocks}.
Cooperative behaviors like \textit{ball\_pass} are triggered via behavior arbitration in robot soccer games \cite{pagello1999cooperative}.
The behavior of coupled oscillatory containment that both members synchronously oscillated around the sheep is observed to be more effective in multi-agent shepherd tasks \cite{nalepka2017herd}.
The previous robotic behavior analysis motivates us to define emergent cooperative behavior in active target tracking, which has no previous investigation. 

The primary contributions of this paper are:
\begin{itemize}
    \item We introduce the concept of a \emph{dynamic occlusion map} that maintains stationary occlusion objects (SOO). 
    UAVs simultaneously detect, track, and share information of OOIs and SOOs with different motion hypotheses. 
    In the conditions of no prior semantic information on occlusions, the dynamic occlusion map allows the DTT to reduce the tracking deterioration by increasing the awareness of the ``just detected'' occlusions in robot motion planning.
    \item Explicit definition of \emph{cooperative behaviors} that we demonstrate as emergent behavior as a consequence of active target tracking. 
\end{itemize}

\section{Active Sensing in Multi-sensor Multi-target Tracking with Unknown Occlusions}

\label{sec:active-sensing}
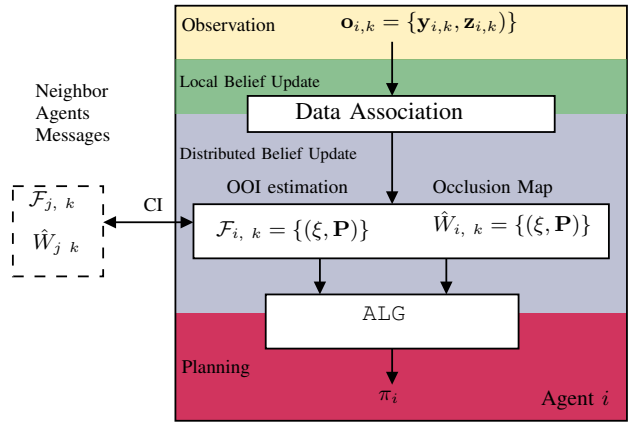
\begin{figure}

\centering
\captionsetup{aboveskip=-5pt}
\tikzset{every picture/.style={line width=0.75pt}} %set default line width to 0.75pt        
\resizebox{1.0\columnwidth}{!}{
\tikzset{every picture/.style={line width=0.75pt}} %set default line width to 0.75pt        

\begin{tikzpicture}[x=0.75pt,y=0.75pt,yscale=-1,xscale=1]

%Shape: Rectangle [id:dp6574513612938255] 
\draw  [draw opacity=0][fill={rgb, 255:red, 209; green, 52; blue, 94 }  ,fill opacity=1 ][dash pattern={on 0.84pt off 2.51pt}] (121,211) -- (371,211) -- (371,271) -- (121,271) -- cycle ;
%Shape: Rectangle [id:dp16064362791323727] 
\draw  [draw opacity=0][fill={rgb, 255:red, 190; green, 193; blue, 212 }  ,fill opacity=1 ][dash pattern={on 0.84pt off 2.51pt}] (121,101) -- (371,101) -- (371,211) -- (121,211) -- cycle ;
%Shape: Rectangle [id:dp9029419899275111] 
\draw  [draw opacity=0][fill={rgb, 255:red, 255; green, 242; blue, 194 }  ,fill opacity=1 ][dash pattern={on 0.84pt off 2.51pt}] (121,41) -- (371,41) -- (371,71) -- (121,71) -- cycle ;
%Shape: Rectangle [id:dp6513468513840754] 
\draw  [color={rgb, 255:red, 0; green, 0; blue, 0 }  ,draw opacity=1 ] (121,41) -- (371,41) -- (371,271) -- (121,271) -- cycle ;
%Shape: Rectangle [id:dp5870153717098483] 
\draw  [draw opacity=0][fill={rgb, 255:red, 63; green, 150; blue, 73 }  ,fill opacity=0.6 ][dash pattern={on 0.84pt off 2.51pt}] (121,71) -- (371,71) -- (371,101) -- (121,101) -- cycle ;
%Straight Lines [id:da15341330680173138] 
\draw    (84,161) -- (128,161) ;
\draw [shift={(131,161)}, rotate = 180] [fill={rgb, 255:red, 0; green, 0; blue, 0 }  ][line width=0.08]  [draw opacity=0] (7.14,-3.43) -- (0,0) -- (7.14,3.43) -- cycle    ;
\draw [shift={(81,161)}, rotate = 0] [fill={rgb, 255:red, 0; green, 0; blue, 0 }  ][line width=0.08]  [draw opacity=0] (7.14,-3.43) -- (0,0) -- (7.14,3.43) -- cycle    ;
%Shape: Rectangle [id:dp706050279432624] 
\draw  [fill={rgb, 255:red, 255; green, 255; blue, 255 }  ,fill opacity=1 ] (131,151) -- (361,151) -- (361,181) -- (131,181) -- cycle ;
%Shape: Rectangle [id:dp7160864047047519] 
\draw  [color={rgb, 255:red, 0; green, 0; blue, 0 }  ,draw opacity=1 ][fill={rgb, 255:red, 255; green, 255; blue, 255 }  ,fill opacity=1 ] (171,201) -- (311,201) -- (311,231) -- (171,231) -- cycle ;
%Shape: Rectangle [id:dp5292480609242216] 
\draw  [fill={rgb, 255:red, 255; green, 255; blue, 255 }  ,fill opacity=1 ] (161,91) -- (331,91) -- (331,111) -- (161,111) -- cycle ;
%Straight Lines [id:da13495231542328634] 
\draw    (241,111) -- (241,148) ;
\draw [shift={(241,151)}, rotate = 270] [fill={rgb, 255:red, 0; green, 0; blue, 0 }  ][line width=0.08]  [draw opacity=0] (7.14,-3.43) -- (0,0) -- (7.14,3.43) -- cycle    ;
%Straight Lines [id:da978839918690734] 
\draw    (241,61) -- (241,88) ;
\draw [shift={(241,91)}, rotate = 270] [fill={rgb, 255:red, 0; green, 0; blue, 0 }  ][line width=0.08]  [draw opacity=0] (7.14,-3.43) -- (0,0) -- (7.14,3.43) -- cycle    ;
%Straight Lines [id:da6705880703861473] 
\draw    (201,181) -- (201,198) ;
\draw [shift={(201,201)}, rotate = 270] [fill={rgb, 255:red, 0; green, 0; blue, 0 }  ][line width=0.08]  [draw opacity=0] (7.14,-3.43) -- (0,0) -- (7.14,3.43) -- cycle    ;
%Straight Lines [id:da7326749463942777] 
\draw    (271,181) -- (271,198) ;
\draw [shift={(271,201)}, rotate = 270] [fill={rgb, 255:red, 0; green, 0; blue, 0 }  ][line width=0.08]  [draw opacity=0] (7.14,-3.43) -- (0,0) -- (7.14,3.43) -- cycle    ;
%Straight Lines [id:da3710714945441367] 
\draw    (241,231) -- (241,248) ;
\draw [shift={(241,251)}, rotate = 270] [fill={rgb, 255:red, 0; green, 0; blue, 0 }  ][line width=0.08]  [draw opacity=0] (7.14,-3.43) -- (0,0) -- (7.14,3.43) -- cycle    ;
%Shape: Rectangle [id:dp8811595463153159] 
\draw  [color={rgb, 255:red, 0; green, 0; blue, 0 }  ,draw opacity=1 ][dash pattern={on 4.5pt off 4.5pt}] (31,141) -- (81,141) -- (81,191) -- (31,191) -- cycle ;

% Text Node
\draw (142,158) node [anchor=north west][inner sep=0.75pt]  [font=\small] [align=left] {$\displaystyle \mathcal{F}_{i,\ k} =\{( \xi ,\mathbf{P})\}$};
% Text Node
\draw (211,43) node [anchor=north west][inner sep=0.75pt]  [font=\small] [align=left] {$\displaystyle \mathbf{o}_{i, k} =\{\mathbf{y}_{i, k} ,\mathbf{z}_{i, k})\}$};
% Text Node
\draw (186,93) node [anchor=north west][inner sep=0.75pt]   [align=left] {Data Association};
% Text Node
\draw (262,153) node [anchor=north west][inner sep=0.75pt]  [font=\small] [align=left] {$\displaystyle \hat{W}_{i,\ k} =\{( \xi ,\mathbf{P})\}$};
% Text Node
\draw (148,136) node [anchor=north west][inner sep=0.75pt]  [font=\footnotesize] [align=left] {OOI estimation};
% Text Node
\draw (262,136) node [anchor=north west][inner sep=0.75pt]  [font=\footnotesize] [align=left] {Occlusion Map};
% Text Node
\draw (123,46) node [anchor=north west][inner sep=0.75pt]  [font=\footnotesize] [align=left] {Observation};
% Text Node
\draw (38.4,142.2) node [anchor=north west][inner sep=0.75pt]  [font=\small] [align=left] {$\displaystyle \mathcal{F}_{j,\ k}$};
% Text Node
\draw (223,206) node [anchor=north west][inner sep=0.75pt]  [font=\normalsize] [align=left] {{\fontfamily{pcr}\selectfont ALG}};
% Text Node
\draw (123,236) node [anchor=north west][inner sep=0.75pt]  [font=\footnotesize] [align=left] {Planning};
% Text Node
\draw (102,146) node [anchor=north west][inner sep=0.75pt]  [font=\footnotesize] [align=left] {CI};
% Text Node
\draw (232,253) node [anchor=north west][inner sep=0.75pt]  [font=\small] [align=left] {$\displaystyle \pi _{i}$};
% Text Node
\draw (322,253) node [anchor=north west][inner sep=0.75pt]  [font=\small] [align=left] {Agent $\displaystyle i$};
% Text Node
\draw (42,83) node [anchor=north west][inner sep=0.75pt]  [font=\footnotesize] [align=left] {Neighbor \\Agents\\Messages};
% Text Node
\draw (122,78) node [anchor=north west][inner sep=0.75pt]  [font=\scriptsize] [align=left] {Local Belief Update};
% Text Node
\draw (122,118) node [anchor=north west][inner sep=0.75pt]  [font=\scriptsize] [align=left] {Distributed Belief Update};
% Text Node
\draw (40.4,163) node [anchor=north west][inner sep=0.75pt]  [font=\small] [align=left] {$\displaystyle \hat{W}_{j\ k}$};

\end{tikzpicture}
}
\vspace{10pt}
\caption{
An occlusion-aware active sensing in DTT paradigm with dynamic occlusion map $\hat{W}$.
}
\label{fig:paradigm}
\end{figure}

In this paper, 
we use the letter $i, j$ as the index of agents; 
$h, k$ as the discrete-time index; 
$N$ as the total number of agents in the team;
integer set $[N] = \{1, 2, ..., N\}$;
words \textit{agent} and \textit{robot} are used interchangeably as the meaning of the decision-making unit in the MRS.
The cooperative $N$-robot target tracking is fit into a POMDP framework as a tuple $\mathcal{P} = (\mathcal{S, A, T, O}, C, \Omega, \gamma)$ similar to \cite{bandyopadhyay2009motion, li2022sma, li2021data}.
Additionally, the horizon $H$ is often considered in decision-making.
The scenario of target tracking is the following: 
Total $N$ UAVs fly at a fixed altitude and track OOIs on a plain 2-dimensional closed AOI; each UAV is equipped with a nadir camera facing $90 \degree$ downward to the ground plane of AOI.
The MRS has a list of semantic labels for OOI and SOO, for instance, $\mathcal{L}_\text{OOI} = \{\text{human}\}, \mathcal{L}_\text{SOO} = \{\text{tree}\}$.
We assume the environment contains a list of $n$ SOOs $W = \{w_1, ..., w_n\}$, with each element existing in the closed area of the AOI.

% Each UAV detects only the OOIs in its local FoV. 
We assume the team contains ideal connectivity for information sharing between any pair of UAVs.
Every agent runs JPDAF with linear Kalman filter to process the local observation and fuses information by the \textit{consensus on information} (CI) algorithm to maintain the global DTT described in \cite{li2021optimizing}.

The \textbf{state} space $\mathcal{X} = (\mathcal{S} \times \chi \times \mathcal{F})$ contains: 
 the joint robots state $s \in \mathcal{S}$  where $\mathcal{S} = \Pi_{i = 1}^N \mathcal{S}_i$. %The robot $i\in[N]$
 and the $i$th individual robot state,  $s_i \in \mathcal{S}_i$, is defined as $s_{i} = (p^{x}_{i}, p^{y}_{i}, \psi_{i}, v^{x}_{i}, v^{y}_{i}) \in \mathbb{R}^5$ in a 2-dimensional horizontal plane including position $p$, yaw angle $\psi$ and velocity $v$, the sensor's FoV $\phi_{i}(s)$ is the projected sensing area parameterized by the robot state and sensor specifications, thus the team's sensor coverage is described as a tuple of their FoVs, $\phi(s) = (\phi_{1}, ..., \phi_{N})$;
the state of OOIs $\chi$ contains the position and velocity of all OOIs;
the filter (estimation) state $\mathcal{F}$ is the Gaussian distribution $\mathcal{N}(\xi, \mathbf{P})$ maintained by the \textit{consensus-based} DTT algorithm \cite{li2021optimizing}.

The joint \textbf{action} space $\mathcal{A} = \Pi_{i=1}^N \mathcal{A}_i$ describes the horizontal velocity command to the team, with each robot $a_{i} = (a^x_i, a^y_i) \in \mathbb{R}^2$ and velocity constraint $|a_i| \leq v_{max, i}$.

The uncertainty in \textbf{observation} brings the observation space $\Omega$ and observation rule $\mathcal{O}$ in POMDP: 
sensors obtains observation $\mathbf{o} = (\mathbf{y}, \mathbf{z}) \in \Omega$ which contains the semantic label $\mathbf{y}$ and localization of the objects $\mathbf{z}$ in sensors' FoVs.
The sensor model assumes uncertainty only in the localization part $\mathbf{z}$. 
For a pair of object-observer $(t, i)$, we use the letter $t$ to represent the index of an OOI, and $i$ is the index of the robot.
The observation law $\mathcal{O}(x|o)$ is the two-fold conditional probability in OOI state $\chi$.
\begin{enumerate}
    \item \textbf{Noise} in localization at time step $k$: 
    \begin{subequations}
    \begin{align}
    \mathbf{z}_{t, i, k} &= \mathbf{H}_k \chi_{t, k} + \mathbf{v}_{t, i, k},  \mathbf{v}_{t, i, k} \sim \mathcal{N}(0, \mathbf{R}_{t, i, k}) \label{eq:observation-model}\\
    \mathbf{R}_{t, i, k} &= \alpha_i \mathbf{G}(\rho_{t, i, k}) 
        \begin{bmatrix}
    0.1 d_{t, i, k} & 0\\
    0 & 0.1\pi d_{t, i, k}
    \end{bmatrix}
    \mathbf{G}(\rho_{t, i, k})^T \label{eq:noise}
    \end{align}
    \end{subequations}
    In \eqref{eq:observation-model}, the localization of $t$ detected by robot $i$ contains a Gaussian white noise $\mathbf{v}_{t, i, k}$.
    The covariance $\mathbf{R}_{t, i, k}$ is the \textit{range-bearing} sensing noise determined by Euclidean distance $d_{t, i, k}$ and bearing difference $\rho_{t, i, k}$ between the sensor and OOI in \eqref{eq:noise}, and scalar $\alpha_i$.
    \item \textbf{Visibility}: the object $t$ is \textit{visible} to robot $i$ at time $k$ if and only if $t$ is inside the FoV and not in any occlusion area; the set of visible objects to robot $i$ is defined
    \begin{equation}
        \label{eq:visibility}
        \mathcal{V}_{i, k} = \{t| (\chi_{t, k} \in \phi_{i, k}) \land (\chi_{t, k} \notin W)  \},
        % \mathcal{V}(t, i) = \left\{ z | \left(\mathbf{z}_{t, i} \in \phi_i(s) \right) \land \left(\mathbf{z}_{t, i} \notin W \right) \right\}
    \end{equation}
and the visible objects to the team is $\mathcal{V}_{k} = \cup_{i \in [N]} V_{i, k}$.
\end{enumerate}

The \textbf{belief-state} space $\mathcal{B(X)}$ depicts the partially observable OOI state $\xi$ by the probability distribution in target tracking problem \cite{li2022sma}.
The belief-state $b_k(x) := b(x_k) \in \mathcal{B(X)}$ contains the belief of every components in the state $b_k(x) = P_{x_k}(x| Z_0, Z_1, ..., Z_k, u_0, ..., u_k) = (b^s_k, b^\chi_k, b^\xi_k, b^P_k)$:
observable state includes robot state $s$ and OOI trajectory's filter state $\mathcal{F}$ ($b^s_k, b^\xi_k, b^P_k$) can be represented by Dirac delta function, e.g. $b^s_k = \delta(s - s_k)$;
the target belief-state is the conditional probability distribution represented by the filter state $F_k$, which is only true for this specific implementation.
% represented by the filter's posterior estimate $b^\chi_k \sim \mathcal{N} (\xi_{k|k}, \mathbf{P}_{k|k})$. 

The belief-state \textbf{transition} $b_{k+1} \sim \mathcal{T}(\cdot | b_k, a_k)$ consists of components corresponding to the three belief-state elements:
i) deterministic velocity command $a_k$ to robot state $b^s_k$;
ii) OOI state $\chi_k$ is stochastic, independent of sensor state $s_k$ and control variable $a_k$;
iii) \textit{nearly constant velocity (NCV)} motion model in filter's trajectory estimation: 
for one OOI with state $\xi_{t, k} = (p_{t, k}^x, p_{t, k}^y, v_{t, k}^x, v_{t, k}^y) \in \mathbb{R}^4$, its dynamics assumption is
\begin{equation}
    \label{eq:dynamics}
    \xi_{t, k + 1} = \mathbf{F}_k\xi_{t, k} + \mathbf{w}_k, \mathbf{w}_k \sim \mathcal{N}(\mathbf{0},\mathbf{Q}).
\end{equation}
% Linear Kalman Filter updates $\mathcal{F}$ based on observation $o_k$.
We use linear Kalman filter and NCV motion model to update the estimated trajectories with semantics in $\mathcal{L}_\text{OOI}$.

% Various objective functions are studied in the planning of active sensing.
% From the tracking side, the information age of all targets is minimized in \cite{jakob2010occlusion}.
% The viewpoint of the camera's viewing angle, OOI projection size, and image space is one sensor-specific objective example \cite{nageli2017real}.
% Mutual information is one desirable objective in applications like SLAM and exploration due to features like submodularity \cite{kantaros2019asymptotically}.
% A practical mixed objective, for instance, combining visibility and power consumption, helps healthcare robot function economically \cite{bandyopadhyay2009motion}.
We consider the cost function $C: \mathcal{B(X)} \times \mathcal{A} \xrightarrow[]{} \mathbb{R}_{\geq 0}$ which represents the total OOIs' expected mean square tracking error. 
The cost function equates to the trace of posterior covariance matrix $\mathbf{P}$ given belief-state transition under Gaussian noise assumption \cite{miller2009coordinated},
\begin{equation}
\label{eq:cost}
\begin{aligned}
C(b_k, a) 
& = \int \mathbb{E}_{w_k, v_{k+1}}
\bigl[ \, ||\chi_{k+1} - \xi_{k+1}||^2 \vert x_k, a
\bigr] \,b^\chi_k(x)dx \\
& = tr[\mathbf{P}_{k+1|k+1}]
\end{aligned}
\end{equation}

A \textbf{policy} $\pi: \mathcal{B(X)} \rightarrow \mathcal{A}$ generates the action selection given a belief-state.
In this cooperative tracking case, the \textbf{objective} accumulates the cost $r_k = C(b_k, a_k)$ over $H$-step look-ahead receding horizon.
The value of belief-state $b$ with policy $\pi$ is the expected accumulated cost over receding horizon $H$ as the following equation.
\begin{equation}
    \label{eq:J_H} 
    \small
J_H^\pi(b) = \mathbb{E}_{\mathcal{T}(\cdot | b_k, a_k)}\biggl[ \,\sum_{k=0}^{H-1} \gamma^k r_k \bigg\vert a_k = \pi(b_k), b_0=b \,\biggl] 
\end{equation}
% $r_k = C(b_k, a_k)$.
The objective function can be equivalently expressed as $J_H^\pi(b) = J_H(b, \pi_1, ..., \pi_N)$ expanding the policy of every agent.

\section{DTT Planning with Unknown Occlusions}

\begin{figure*}

\centering
\captionsetup{aboveskip=-5pt}
\tikzset{every picture/.style={line width=0.75pt}} %set default line width to 0.75pt        
\resizebox{1.8\columnwidth}{!}{
\input{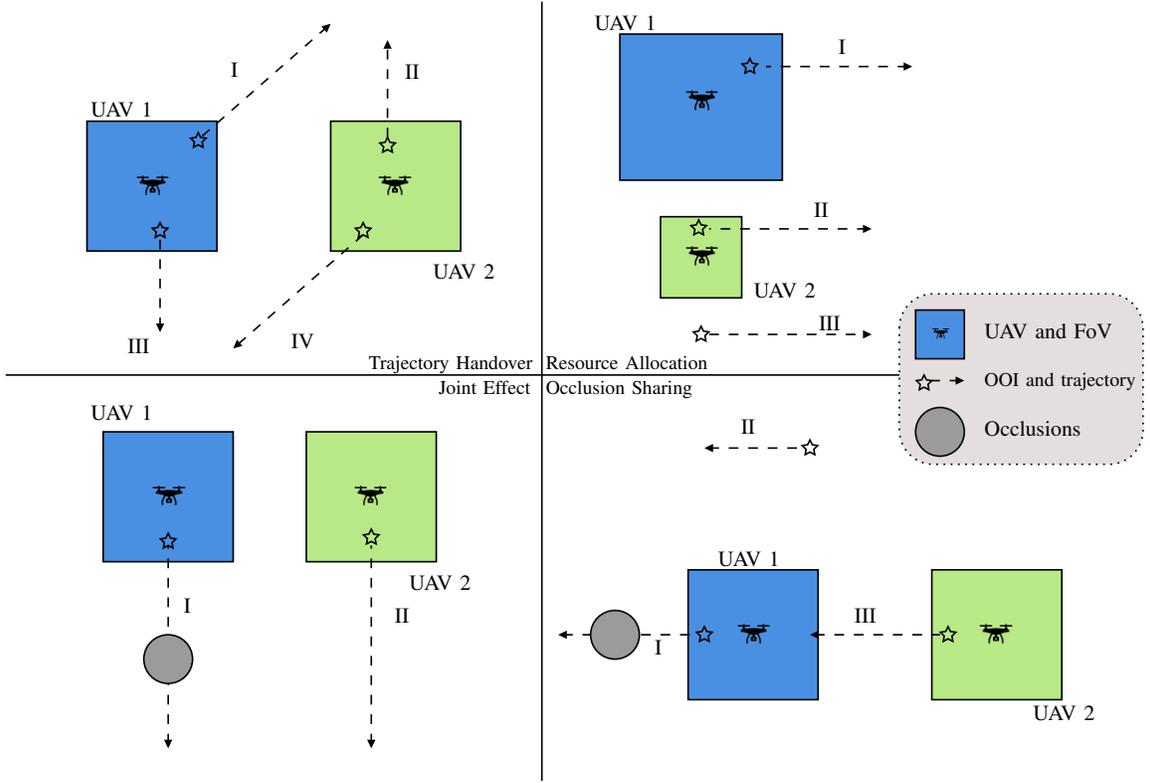}
}
\vspace{10pt}
\caption{
DTT cases with potential cooperative behavior.
In these four cases, the trajectories ground truth are dashed arrows.
All OOIs marked in every scenario are initialized with identical uncertainty matrix $\mathbf{P}$ in trajectory estimation $\mathcal{F}_0$.
The duration, $L$, of each scenario varies.
}
\label{fig:scenarios}
\end{figure*}

\label{sec:coop-algorithms}

We denote a general planning algorithm \texttt{ALG} generates the policy with the objective \eqref{eq:J_H} defined. 
A direct approach is the centralized style \texttt{ALG} to directly solve the optimization problem. 
% The scalability of the team raises the issue of computation and communication.
Pure \textit{decentralized} planning Dec-POMDP removes all dependency of other agents in communication so that every agent $i$ generate an optimal plan or macro-actions \cite{ragi2014decentralized} based on \textit{local trajectory estimation} $\mathcal{F}_{i}$ where 
\begin{equation}
    \label{eq:dec-pomdp}
    \pi^{\text{\tiny Dec-POMDP}}_i \in \argmax_{\pi} J^{\pi}_H(b; \mathcal{F}_{i})
\end{equation}
Applying the consensus algorithm in DTT promises the bound of difference in $\mathcal{F}_{i}$ over agents \cite{li2021optimizing}, which leads the Dec-POMDP solution result to equate to that of a centralized POMDP.

% sequential method
The increase in team size $N$ exponentially increases the computational complexity of \eqref{eq:dec-pomdp}.
A more tractable \texttt{ALG} is preferred from the implementation's aspect, albeit at the expense of suboptimality. 
Here, we engage SMA-NBO, an \textit{agent-by-agent} planning approach based on \textit{intention} \cite{li2022sma}. 
Denote the intention of agent $i$ as $\Bar{\pi}_i$.
SMA-NBO generates a plan, $k = 1$ is $\big (a_1, a_2, .. a_H | a_k = \pi(b_k) \big )$, at every time step.
The initial action $a_1$ is immediately applied as the control, conforming to other receding horizon approaches. However, in next decision-making stage $k = 2$, SMA-NBO exploits the residual action sequence, communicating it to the other fleet members as future \emph{intentions} to inform the next decision epoch, i.e. $\Bar{\pi} \coloneqq (a_2, .. a_H)$.
Policy $\pi^{\text{\tiny SMA}}_i$ is the single agent policy improvement based on $1:i-1$ agents' updated policy $\pi^{\text{\tiny SMA}}_{1:i-1}$ and the intention of the rest $\Bar{\pi}_{i+1:N}$.
\begin{subequations}
\label{eq:agent-by-agent}
\begin{align}
\pi^{\text{\tiny SMA}}_1 &= \argmax_{\pi_1} J_H(b, \pi_1, \Bar{\pi}_{2:N}) \label{eq:agent-by-agent:1}\\
\pi^{\text{\tiny SMA}}_i &= \argmax_{\pi_i} J_H(b, \pi^{\text{\tiny SMA}}_{1:i-1}, \pi_i, \Bar{\pi}_{i+1:N}) \label{eq:agent-by-agent:i}\\
\pi^{\text{\tiny SMA}}_N &= \argmax_{\pi_N} J_H(b, \pi^{\text{\tiny SMA}}_{1:N-1}, \pi_N)\label{eq:agent-by-agent:n}
\end{align}
\end{subequations}

% \begin{figure}

% \centering
% \captionsetup{aboveskip=-5pt}
% \tikzset{every picture/.style={line width=0.75pt}} %set default line width to 0.75pt        
% \resizebox{1.0\columnwidth}{!}{
% \input{pics/paradigm.tikz}
% }
% \vspace{10pt}
% \caption{
% An occlusion-aware active sensing in DTT paradigm with dynamic occlusion map $\hat{W}$.
% }
% \label{fig:paradigm}
% \end{figure}

\subsection{Unknown Occlusions}
\label{sec:occlusion}

As stated in Section \ref{sec:active-sensing}, an occlusion-aware planning requires planner to check visibility of all OOI-observer pairs, $(t, i)$, over look-ahead horizon.
In this specific problem, a valid \`{a} priori occlusion map $W$ contributes to $J_H^{\pi}$ in covariance matrix update of information filter format \eqref{eq:cost}:
\begin{equation}
    \mathbf{P}_{k+1|k+1, t} = \big (\mathbf{P}_{k+1|k, t} + \sum_{t \in \mathcal{V}_k} \mathbf{H}_k^T \mathbf{R}_{t, i, k}^{-1}  \mathbf{H}_k \big ) ^ {-1}
\end{equation}
If the \`{a} priori map $W$ is invalid or not available, the behavior of \texttt{ALG} is no longer occlusion-aware.

However, unknown SOOs, if recognized through observations, are able to be maintained in DTT in the paradigm of Figure~\ref{fig:paradigm}.
Here we assume the observation law is consistent in sensing noise model \eqref{eq:noise} and visibility \eqref{eq:visibility} for both OOI and SOO.
At each time step, the data association algorithm JPDAF updates local observation tuples of label and position $(\mathbf{y}_k, \mathbf{z}_k)$.
A semantic-aware motion model can be applied to SOOs \eqref{eq:dynamics}, for example, static assumption for \textit{tree} object is $\mathbf{F}_k = \mathbf{I}, \mathbf{Q} = \mathbf{0}$.
The estimation of SOOs forms the \textit{dynamic occlusion map} $\hat{W}_{i, k}$, which is shared and updated by CI.
DTT framework makes the temporal knowledge of global SOOs accessible to all members of MRS.
The updated map $\hat{W}_{i, k}$ is immediately exploited in \eqref{eq:visibility} to approximate $W$ in planning \texttt{ALG}.
While the dynamic occlusion map only contains detected estimations of SOO, it increases the occlusion-aware in near future steps if an OOI is moving towards an SOO estimation.
Also, the sharing of SOO estimation increases the cooperation in MRS, which is analyzed in Section \ref{sec:behavior-occlusion}.

\section{Emergent Behavioral Analysis in Active Sensing}

\label{sec:emergent-behavior}

In this section, we define the emergent cooperative behaviors in DTT task.
We generalize the scenarios in DTT in Figure~\ref{fig:scenarios}.
The context of the scenarios includes two cases without occlusion (Trajectory Handover and Resource Allocation), and two cases with occlusions (Joint Effect and Occlusion Sharing), illuminating emergent cooperation in active target tracking.
Simulations are run in all scenarios with the following setup.

\subsection{Simulation Setup}

The primary purpose of the simulation is to provide empirical statistics for emergent behavioral analysis.
All numerical simulations are conducted in advanced computing resources provided by Texas A\&M University High Performance Research Computing \footnote{TAMU HPRC: \url{https://hprc.tamu.edu/}}.

The planning \texttt{ALG} SMA-NBO and Dec-POMDP with NBO are implemented to compare and contrast their cooperative tendencies even though their common objective function does not explicitly encode a requirement for task collaboration.
Both \texttt{ALG}s use the nominal belief trajectory approaches in cost function approximation over trajectory belief state \eqref{eq:cost}, which predicts the OOI trajectories by motion model \eqref{eq:dynamics} without noise \cite{li2022sma}.
This nominal belief trajectory is regarded as the maximum likelihood trajectory under the Gaussian assumptions present in the NCV model.
Particle swarm optimization (PSO) is implemented in both \texttt{ALG}s' to numerically optimize the objective function \ref{eq:cost}.

In Trajectory Handover and Resource Allocation, we set the look-ahead horizon of \texttt{ALG} to $H=$ 1 and 5.
The $H=1$ horizon provides the \textit{baseline} greedy solution in behavior analysis.
Both scenarios are repeated in 40 trials with each combination of \texttt{ALG} and $H$.

In the active sensing paradigm of Figure~\ref{fig:paradigm}, the occlusion map $W$ is the crucial component to trigger the occlusion-related behaviors. 
The Joint Effect and Occlusion Sharing scenarios exemplify said behaviors. 
The \texttt{ALG} is fixed with SMA-NBO and $H = 5$, and simulations are repeated 40 times over the different occlusion map setups: \`{a} priori occlusion map, dynamic occlusion map and no occlusion map.
We investigate the efficacy of the dynamic occlusion map from the behavioral perspective.

\subsection{Cooperative Behavior in DTT}
% reasoning on the scenarios, cooperation
\label{sec:cooperative}

MRS's redundancy and heterogeneity make cooperative behaviors in active sensing possible.
Different from exploration-oriented tasks, target tracking objectives inspire OOI-centric actions which strive for increased target visibility opportunities to minimize the localization uncertainty of OOI estimations.
Multiple divergent OOI trajectories challenge this ``eyes-on" behavior for single robot cases. 
The limited FOV requires a balance between visibility over the set of trajectories, which can be discerned from  \eqref{eq:cost}.
In DTT, robots overcome this challenge by distributing the team member responsibilities over OOIs estimations in requisite directions.
To recognize this responsibility, we define \textit{m-Ownership} in analyzing the behavior of MRS over the simulation horizon.

\begin{definition}[\ownership$(i, t, k)$]
Robot $i$ obtains the \textit{m-Ownership} of the target $t$ at time step $k$ if robot $i$ detects $t$ for $m$ consecutive time steps, i.e., $\forall 0 \leq h \leq m, t \in \mathcal{V}_{i, k + h}$.
\end{definition}
Let $M_k = (M_{1, k}, ..., M_{N, k})$ represent the \ownership\ of all $N$ robots at time index $k$, with every subset $M_{i, k}$ containing the set of target indices ``\textit{m\_owned}" by robot $i$.
Then we define the behavior of reassignment of ownership.
% \begin{definition}[\textit{Ownership\_Change}
% $(i, k_0) \xrightarrow{t} (j, k_L)$]
\begin{definition}
[\bhonwershipchange{i}{j}{t}]
The behavior \textit{Ownership\_Change} happens between robot $i, j$: at time step $0$ target $t \in M_{i, k} \land t \notin M_{j, k}$, and after $L$ steps the ownership goes to $j$, i.e., $t \in M_{j, k + L} \land t \notin M_{i, k + L}$.
\end{definition}
\ownershipchange\ depicts the visibility change of target $t$ between a pair of robot $i, j$.
One example is the Trajectory Handover scenario in Figure~\ref{fig:scenarios}.
At time $k=0$, the \ownership\ $M_0 = \big( M_{\text{\scriptsize UAV1}, 0}:(\text{\small I, III}), M_{\text{\scriptsize UAV2}, 0}:(\text{\small II, IV}) \big)$.
The trajectory estimation $\mathcal{F}_0$ predicts the divergence of trajectories owned by both UAVs (I and II, III and IV).
After $L$ time steps when trajectories owned by the same UAV are sufficiently separated (in terms of distance), there is intuition that the optimal cooperative outcome should leverage a "zone defense" strategy. 
More concisely, the desired resulting ownership is $M_L^* = \big( (\text{\small I, II}), (\text{\small III, IV}) \big)$, where the robot is not specified since both permutations of the robot index is of the equivalent level of cooperation.
From $M_0$ to $M_L^*$, the \ownershipchange\ happens in one of the two cases below.
\begin{align*}
    % \centering
    (0, \text{\small UAV1} \xrightarrow{\text{\small I}, L} \text{\small UAV2}) &\land 
    (0, \text{\small UAV2} \xrightarrow{\text{\small IV}, L} \text{\small UAV1})\\
    (0, \text{\small UAV1} \xrightarrow{\text{\small III}, L} \text{\small UAV2}) &\land 
    (0, \text{\small UAV2} \xrightarrow{\text{\small II}, L} \text{\small UAV1})
\end{align*}

\begin{figure}
    \centering
    \captionsetup{aboveskip=-5pt}
\tikzset{every picture/.style={line width=0.75pt}} %set default line width to 0.75pt        
\resizebox{1.0\columnwidth}{!}{
\input{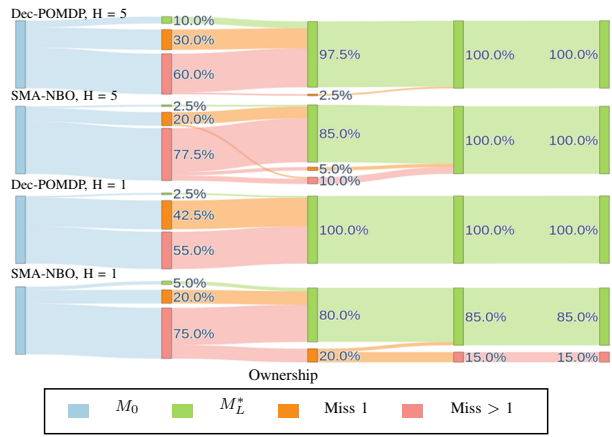}
}
\vspace{10pt}
    \caption{The Sankey diagram of OOI ownership in Trajectory Handover scenario.}
    \label{fig:sankey_trajhandover}
\end{figure}

\textit{Simulation analysis of Trajectory Handover}:
The \ownership[5]\ over the simulation time is plotted in Figure~\ref{fig:sankey_trajhandover}.
In each Sankey diagram, the \ownership[5]\ state over simulation time is visualized: from left to right, each column represents the frequency of \ownership[5]\ at time step $k = 0, \frac{L}{4}, \frac{L}{2}, \frac{3L}{4}, L$ over repeated trials.
The legend ``Miss 1" and ``Miss $>$ 1" refer to the ownership conditions that any UAV, respectively, does not own one or more than 1 OOIs.
In Trajectory Handover, UAVs have trouble maintaining the original \ownership[5]\ at $k = \frac{L}{4}$. 
The \ownershipchange\ behavior is observed from $k=0$ to $k =\frac{L}{2}$ or $\frac{3L}{4}$ with UAVs temporally leaving some OOI trajectories outside of FoVs (the orange and red nodes at $k = \frac{L}{4}$ and $\frac{L}{2}$).
Comparing Sankey diagram between different \texttt{ALG}s, the Dec-POMDP is generally more cooperative.
In $H=5$ case, both Dec-POMDP and SMA-NBO achieve 100\% $M^*_L$, but Dec-POMDP has a higher frequency of performing $M^*_L$ at early stages, for example, 97.5\% v.s. 85 \% at half-time.
SMA-NBO sees more significant improvement by increasing the horizon since  15 \% of simulation trials of myopic SMA-NBO ($H = 1$) end at less cooperative \ownership[5].
The analysis of \ownershipchange\ behavior confirms the slightly reduced optimality of the computationally-efficient SMA instantiation of NBO compared to Dec-POMDP. 
Furthermore, the contrast between the greedy ($H=1$) and the non-myopic ($H=5$) behaviors indicates the cooperative benefits implicitly extracted via look-ahead planning in the Trajectory Handover scenario.

\begin{figure}
    \centering
    \captionsetup{aboveskip=-5pt}
\tikzset{every picture/.style={line width=0.75pt}} %set default line width to 0.75pt        
\resizebox{1.0\columnwidth}{!}{
\input{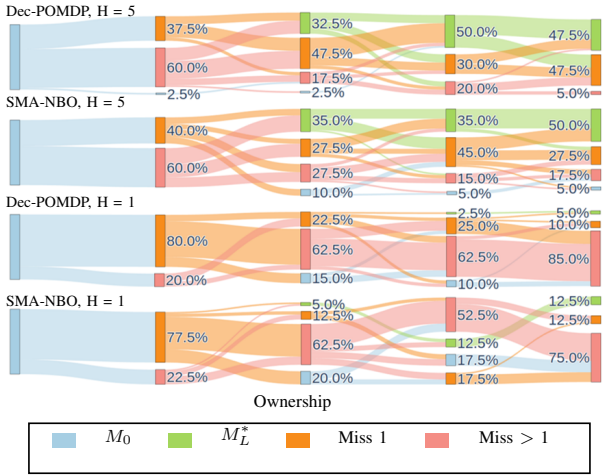}
}
\vspace{10pt}
    \caption{The Sankey diagram of OOI ownership in Resource Allocation scenario.}
    \label{fig:sankey_resource_allocate}
\end{figure}

Beyond tracking divergent trajectories, the cooperative \ownershipchange\ can also occur because of MRS heterogeneity.
The Resource Allocation scenario in Figure~\ref{fig:scenarios} contains the UAVs with different FoV sizes, and neither of the UAVs has visibility of target III.
If keeping the same ownership, UAV 2 needs to move in a zig-zag pattern to track trajectories II and III, also reported in Fig 1 of \cite{miller2009coordinated}.
With the exchange of position, the optimal cooperative behavior is $M_L^* = \big( M_{\text{\scriptsize UAV1}, L}:(\text{\small I}), M_{\text{\scriptsize UAV2}, L}:(\text{\small II, III}) \big)$.
The allocation of FoV resources to $M^*_L$ is cooperative since the visibility of OOIs is improved.

\textit{Simulation analysis of Resource Allocation}:
Similar to Figure~\ref{fig:sankey_trajhandover}, the \ownership[3]\ of repeated simulations are shown in Figure~\ref{fig:sankey_resource_allocate}.
The \ownershipchange\ behavior is more challenging in this scenario than Trajectory Handover; only around half of the simulations achieve $M^*_L$ with horizon $H=5$.
The \ownership[3]\ flow of Dec-POMDP and SMA-NBO at the same horizon has a close probability of achieving the cooperative state.
Given greedy horizon, $H=1$, both planners in the bottom two Sankey diagrams have the rare chance (5 \% in Dec-POMDP and 12.5 \% in SMA-NBO) to allocate the resource.
Unlike Trajectory Handover that from $M_0$ to $M^*_L$ every UAV owns one unchanged target, in Resource Allocation, the \ownership[3]\ of both UAVs from $M_0$ to $M^*_L$ is entirely different.
The condition of achieving the cooperative \ownershipchange\ requires sufficient look-ahead horizon and avoiding all local optima, which is a stringent task for a numerical approach. Thus, the cooperative performance in this scenario is not as obvious.  However, the longer lookahead horizons exhibit cooperative tendencies.  

\subsection{Cooperative DTT with Occlusions}

\label{sec:behavior-occlusion}

The awareness of occlusion provides unique emergent cooperative behavior in DTT tasks.
A general reasoning of occlusion-aware is that if one trajectory $t$ will be occluded in the future time interval $[h, L-h]$, the sensing resource should be allocated for other trajectories during this interval.
\begin{definition}[\textit{Occlusion\_Aware}$(k_0, t, W, L, h)$]
The OOI $t$ is occluded at all time steps $k \in [k_0 + h, k_0 + L-h], \chi_{t, k} \in W$.
We say the behavior is \textit{Occlusion\_Aware} with the satisfaction of all three conditions below.
\begin{enumerate} 
    \item $\exists S_1\subseteq [N]$, $S_1 \neq \emptyset$ s.t. $\forall i \in S_1$, $t \in M_{k_0, i}$;
    \item $\nexists i \in [N]$ s.t. $\forall k \in [k_0 + h, k_0 + L-h]$, $(M_{k, i} = \emptyset) \land (\chi_{t, k} \in \phi_{i, k})$;
    \item $\exists S_2\subseteq [N]$, $S_2 \neq \emptyset$ s.t. $\forall i \in S_2$, $t \in M_{k_0 + L, i}$;
\end{enumerate}
\end{definition}
In the definition of \occlusionaware\, the first and third conditions make sure OOI $t$ is visible to at least one robot before and after the occluded period $[k_0 + h, k_0 + L - h]$;
the second condition states that when $t$ is occluded, all robots should find other OOIs rather than desperately waiting for $t$ if $t$ is the \textit{sole} OOI inside its FoV, which reflects the awareness of occlusion.

One example is the Joint Effect scenario in Figure~\ref{fig:scenarios}.
Trajectory I goes under the occlusion in time interval $[h, L-h]$, and the \ownership\ at initial position is $M_0 = \big( M_{\text{\scriptsize UAV1}, 0}:(\text{\small I}), M_{\text{\scriptsize UAV2}, 0}:(\text{\small II}) \big)$.
The cooperative behavior of \textit{Occlusion\_Aware} here expects that trajectory II should be jointly covered by both UAVs until trajectory I comes outside of the occlusion, i.e. $\forall k \in [h, L-h], M_k = \big( M_{\text{\scriptsize UAV1}, k}:(\text{\small II}), M_{\text{\scriptsize UAV2}, k}:(\text{\small II}) \big)$.
At the ending stage, one UAV covers trajectory I, $M_L = \big( (\text{\small I}), (\text{\small II}) \big)$.

\begin{figure}
    \centering
    \includegraphics[width=.49\textwidth]{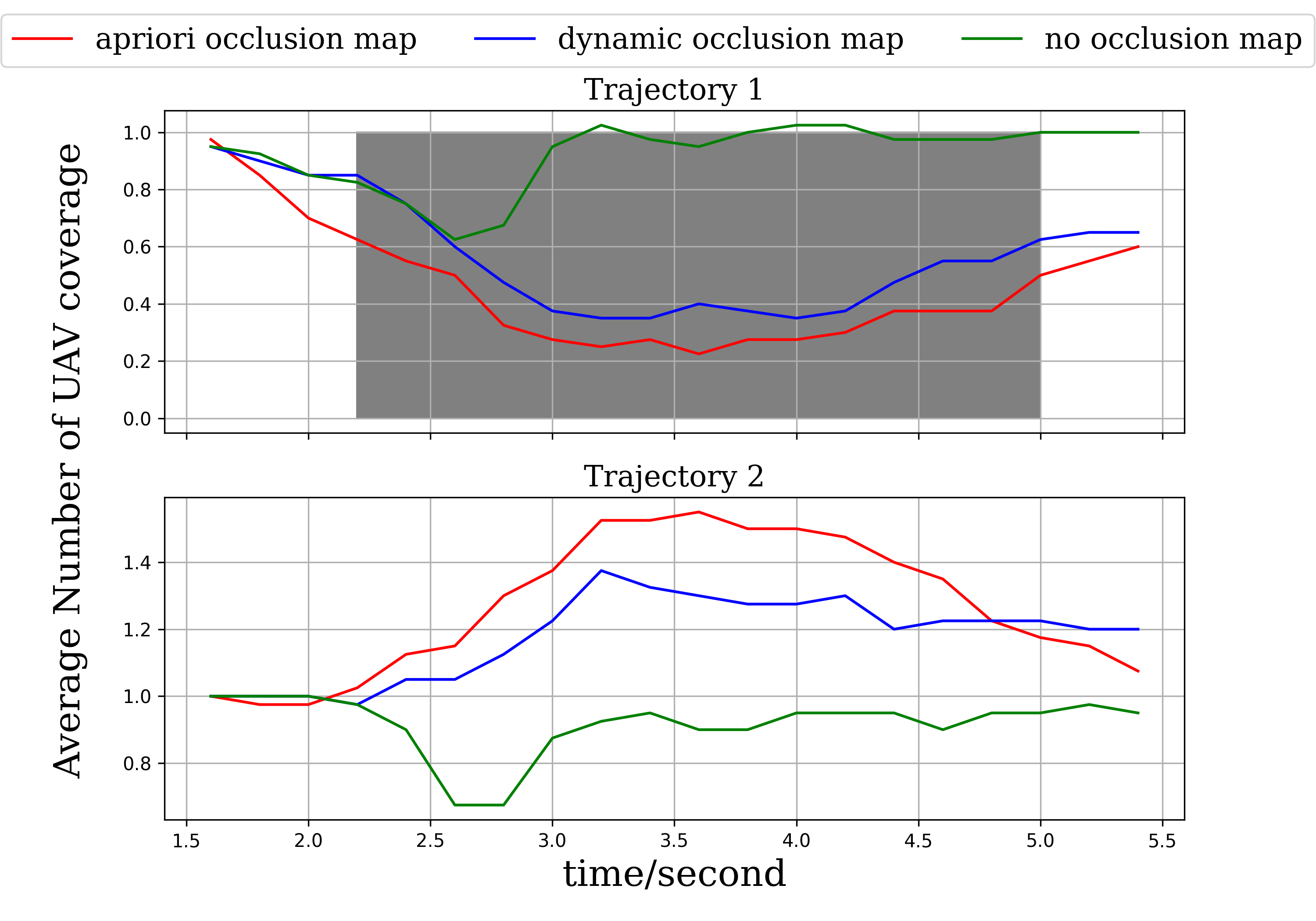}
    \caption{The trajectory coverage for Joint Effect scenario. 
    The grey area indicates this trajectory is under occlusion area.}
    \label{fig:jointeffect-analysis}
\end{figure}

\textit{Simulation analysis of Joint Effect}:
Figure~\ref{fig:jointeffect-analysis} contains the simulation results of the Joint Effect. 
The average number of UAV coverage (\textit{1-Ownership}) is plotted over the simulation horizon per trajectory.
The grey area between 2.2 $\sim$ 5.0 second is the time interval that trajectory I is under occlusion.
From this figure, we compare the behavior of SMA-NBO with different occlusion map knowledge: \`{a} priori, dynamic, and no occlusion map.
The behavior of no occlusion map is very distinguishable since it still continuously attempts to observe trajectory I during the occluded period.
The \`{a} priori and dynamic maps significantly decrease trajectory I's converge at the grey area and, simultaneously,  more than one UAV's coverage over trajectory II.
The empirical result indicates the \occlusionaware\ behavior of \planalg\ with \`{a} priori and dynamic occlusion map.
However, there is a notable change delay at the blue line of 1.6 $\sim$ 2.2 second compared to the red line.
The dynamic map requires the detection of occlusion before the generation of occlusion-aware planning, which explains the slower response of the blue line.

Dynamic occlusion map works based on exploring the environment, which contains a specific form of cooperative behavior.
Once a new SOO is detected by robot $i$, this local information is shared with all robots in the MRS, illustrated in Figure~\ref{fig:paradigm}.
This knowledge sharing is a cooperative behavior which increases the \textit{awareness of SSOs} over the team.

\begin{definition}[\textit{Occlusion\_Sharing}]
Define an \textit{occlusion-detection event} $e$ as an occlusion $w \in W$ is detected by robot $i$ at time step $k$, i.e. $e = \{w \in \phi_{i, k}\}$, then $\forall h > k$ and any trajectory $t$.

The definition of \occlusionshare\ is loosely based on the probability of behavioral event \occlusionaware\ in the future, 
\begin{equation}
    \label{eq:occlusion-share}
    \mathbb{P}[\textit{OA}(t, W_h, L, h) | e] \geq \mathbb{P}[\textit{OA}(t, W_h, L, h)],
\end{equation}
where the term \textit{OA} is the abbreviation of \occlusionaware. 
\end{definition}
We use the last scenario, Occlusion Sharing, in Figure~\ref{fig:scenarios} to demonstrate this concept.
The three trajectories and two UAVs are initiated with trajectory II not owned initially.
If following trajectory I, the UAV 1 will detect the only occlusion at time $k_1$.
Later at time $k_2 > k_1$, trajectory III will go inside the same occlusion.
If this occlusion information is shared, we expect a higher probability of \occlusionaware\ behavior of trajectory III.

\begin{figure}
    \centering
    \includegraphics[width=.48\textwidth]{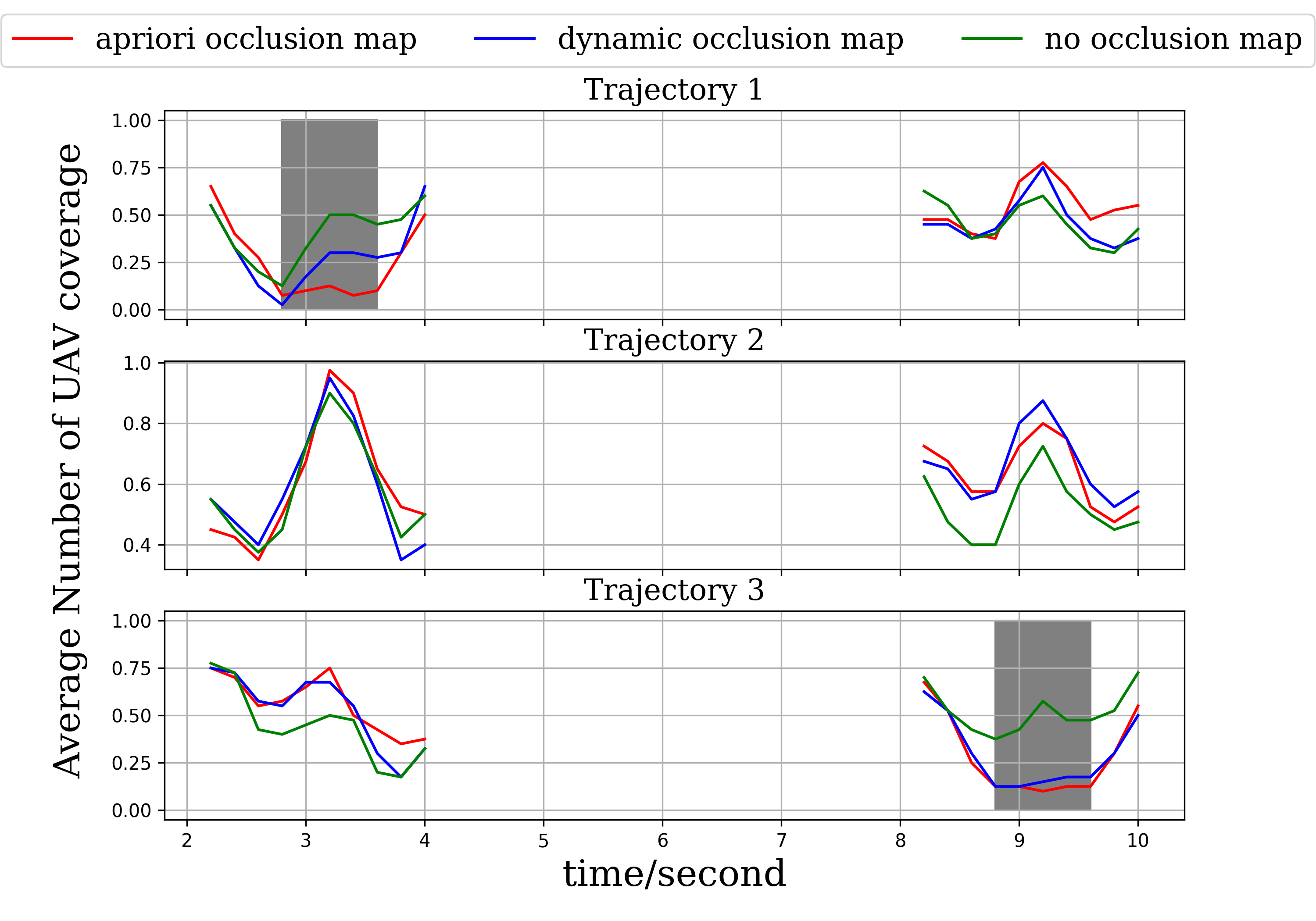}
    \caption{The trajectory coverage for Occlusion Sharing scenario. The grey area indicates this trajectory is under occlusion area.}
    \label{fig:occlusion-aware-analysis}
\end{figure}
\textit{Simulation analysis of Occlusion Sharing}:
The sensing behavior of Occlusion Sharing is visualized in the same way as Joint Effect in Figure~\ref{fig:occlusion-aware-analysis}. 
The \occlusionaware behavior exits in two time intervals: 2.8 $\sim$ 3.6 second over trajectory I, and 8.8 $\sim$ 9.6 second over trajectory III.
In the first occlusion event, we observe the awareness ranking in the order of effectiveness as \`{a} priori, dynamic and no occlusion map based on the average number of sensor FoV's overlapping trajectory I during the grey interval.
Since trajectory II is not covered initially, there is an increase of coverage to balance the uncertainty from UAV 1 and 2: blue and red line has less appearance of UAV around occlusion area to cover trajectory II, and green line has less occurrence of sensing over trajectory III.
The empirical result of behavior around first time interval is identical to Joint Effect.
However, during second interval when trajectory III is occluded, the coverage behavior of dynamic and \`{a} priori occlusion map is almost identical. 
The behavior of no occlusion map still wastes sensing effort over the occlusion area, and the blue and red curve is overlapped over all three trajectories.
The close performance of dynamic occlusion map to \`{a} priori occlusion map indicates the increase of \occlusionaware after the detection and sharing of occlusion.
This validates the definition of \occlusionshare.

\section{Conclusion}

This paper defines three types of emergent cooperative behaviors: \ownershipchange, \occlusionaware, and \occlusionshare.
From four DTT scenarios with potential cooperative behaviors, the defined emergent behavior validates 
\begin{itemize}
    \item the cooperativeness of planning algorithms with  differing computational complexities: Dec-POMDP and SMA-NBO;
    \item the increase in cooperative behaviors elicited from non-myopic solution approaches when compared to greedy approaches;
    \item the efficacy of dynamic occlusion map with unknown occlusion;
    \item the improvement of \occlusionaware\ based on the exploration-sharing paradigm of dynamic occlusion map.
\end{itemize}

The scenarios presented herein demonstrate initial results based on their straightforward configurations. 
However, this provides sufficient insight to improve the dynamic occlusion map for more general SOO geometries and investigate the online distributed occlusion fusion approaches.

% future works?
% 1. The fuse of SOO in dynamic occlusion map can be improved by fusing of the geometry/shape configuration of SOOs.\\
% 2. The emergent cooperative behavior defined can be imitated which forms a context-specific behavior-oriented active sensing approach, to be studied in future.

\bibliographystyle{IEEEtran}
\bibliography{IEEEabrv, reference}

\end{document}